\documentclass[runningheads]{llncs}

\usepackage{mystyle}

\begin{document}
\title{Distorted Distributional Policy Evaluation for Offline Reinforcement Learning}
\titlerunning{Distorted DPE for Offline RL}
%
\author{Ryo Iwaki \and
Takayuki Osogami
}
\authorrunning{R. Iwaki and T. Osogami}
\institute{IBM Research - Tokyo, Japan\\
\email{\{Ryo.Iwaki@,OSOGAMI@jp.\}ibm.com}}
\maketitle              
\begin{abstract}

While Distributional Reinforcement Learning (DRL) methods have demonstrated strong performance in online settings, its success in offline scenarios remains limited. We hypothesize that a key limitation of existing offline DRL methods lies in their approach to uniformly underestimate return quantiles. This uniform pessimism can lead to overly conservative value estimates, ultimately hindering generalization and performance. To address this, we introduce a novel concept called quantile distortion, which enables non-uniform pessimism by adjusting the degree of conservatism based on the availability of supporting data. Our approach is grounded in theoretical analysis and empirically validated, demonstrating improved performance over uniform pessimism.

\end{abstract}

\section{Introduction}

\begin{wrapfigure}{r}[0pt]{0.3\linewidth}
  \vspace{-20pt}
  \centering
  \includegraphics[width=0.3\textwidth]{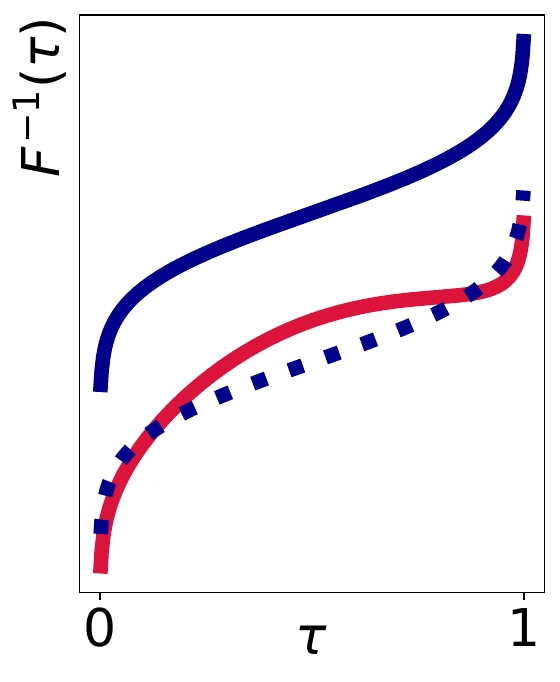}
  \vspace{-20pt}
  \caption{
    Navy: true quantiles.
    Navy(dashed): quantiles w/ uniform pessimism.
    Red: quantiles w/ non-uniform pessimism.
  }
  \label{fig:concept}
  \vspace{-20pt}
\end{wrapfigure}

In reinforcement learning, distributional approaches have shown significant performance gains. Unlike conventional methods that estimate only the expected return, Distributional Reinforcement Learning (DRL) learns the distribution of returns, providing richer information for decision-making \cite{Morimura2010a,Morimura2010b,Bellemare2017}. This distribution awareness not only enables DRL to frequently outperform traditional approaches but also facilitates natural extensions to risk-sensitive settings \cite{Dabney2018IQ,Lim2022}, as exemplified by the discovery of faster matrix multiplication algorithms using a risk-seeking DRL \cite{Fawzi2022}.

In many real-world applications of RL, however, exploration is prohibitively risky or expensive, and constructing reliable simulators is often challenging. Offline RL addresses these limitations by enabling agents to learn from pre-collected datasets, eliminating the need for trial-and-error in real or simulated environments \cite{Levine2020}.

Since the setting of offline RL does not allow an agent to evaluate the quality of actions (or state-action pairs) through direct interaction with the environment, it becomes essential to properly handle out-of-distribution (OOD) actions, which are not well represented in the training data. A widely adopted strategy is the principle of \emph{pessimism} \cite{Buckman2021,Jin2021}, which involves conservatively estimating the expected returns of OOD actions. By assigning lower values to uncertain or unsupported actions, this approach encourages the policy to favor in-distribution actions, often leading to robust and high-performing policies in practice.

Given the demonstrated success in online settings, DRL is a promising candidate for offline RL scenarios as well. However, the challenge of OOD actions becomes even more pronounced in the distributional approach. Offline DRL requires sufficient data to capture the shape of that distribution, but offline datasets often lack the diversity and coverage needed to reliably estimate the distribution, making robust distributional estimation particularly difficult.

Following the principle of pessimism, the quantile values, representing a return distribution, are ``uniformly'' underestimated in existing offline DRL methods \cite{Ma2021} (see the dashed curve in Figure~\ref{fig:concept}). Since the tails are treated equally as the rest of the distribution, this uniform pessimism can lead to over-conservatism, ultimately hindering both generalization and overall performance \cite{Lyu2022Mildly}.

We hypothesize that the tails should be estimated with greater pessimism than other regions, such as the modes. To investigate this, we analyze offline distributional evaluation and introduce a novel evaluation method that applies non-uniform pessimism through a new concept called \emph{quantile distortion} (see the red curve in Figure~\ref{fig:concept}). This approach enables estimating the values in the tails with greater pessimism, while avoiding over pessimism in well-supported regions.  The introduction of non-uniform pessimism for offline DRL constitutes the primary contribution of this paper.

We validate the effectiveness of non-uniform pessimism through experiment on a simulated benchmark and demonstrate its superiority over uniform pessimism,
which marks a significant advancement in offline DRL.
Given that DRL has the potential to outperform other offline RL methods in certain settings, it should be considered as a candidate within the broader framework of automated RL \cite{Franke2021,Marinescu2022}, where the optimal policy is selected from a diverse set of candidate configurations, which may vary in learning algorithms, hyperparameters, model architectures, and other design choices.

\section{Preliminary}

\subsection{Distributional Reinforcement Learning}

We consider an infinite horizon discounted Markov decision process (MDP).
An MDP is specified by a tuple
$(\mathcal{S},\mathcal{A},\mathcal{P},\mathcal{R},\rho_0,\gamma)$,
where $\mathcal{S}$ is a finite set of possible states of an environment,
$\mathcal{A}$ is a finite set of possible actions which an agent can choose,
$\mathcal{P}: \mathcal{S} \times \mathcal{A} \to \mathscr{P}(\mathcal{S})$
is a Markovian state transition probability distribution,
$\mathcal{R}: \mathcal{S} \times \mathcal{A} \to \mathscr{P}(\mathbb{R})$
is an immediate reward probability,
$\rho_{0}: \mathcal{S} \to \mathbb{R}$ is an initial state distribution,
and $\gamma \in (0, 1)$ is a discount factor.
We are interested in evaluating or optimizing a policy
$\pi: \mathcal{S} \to \mathscr{P}(\mathcal{A})$.
For a probability distribution $\nu\in\mathscr{P}(\mathbb{R})$
and its random variable $X$,
the cumulative distribution function (CDF) $F_{\nu}$
and the quantile function (inverse CDF) $F_{\nu}^{-1}$
are given by $
  F_{\nu}(x)
  = \mathbb{P}\left[X \leq x\right]
$ and $
  F_{\nu}^{-1}(\tau)
  = \inf \left\{ x\in\mathbb{R}: \tau \leq F_{X}(x) \right\}
$, respectively.
We also write $F_{X} = F_{\nu}$ and $F_{X}^{-1} = F_{\nu}^{-1}$ when $X\sim\nu$, meaning that $X$ is distributed according to $\nu$.

In distributional RL, the subject of study is the conditional distribution $\eta(s,a)$ of the random return
$
  Z(s,a) = \sum_{t=0}^{\infty} \gamma^{t} r_{t}
$ given $S_{0}=s, A_{0}=a$.
For this conditional distribution $\eta\in\mathscr{P}(\mathbb{R})^{\mathcal{S} \times \mathcal{A}}$,
the following equality characterizes the distributional Bellman operator $\mathcal{T}^\pi$ on CDF:
\begin{align}
  F_{\mathcal{T}^{\pi} \eta (s,a)}(z)
  &=
  \sum_{(s^{\prime}, a^{\prime})\in\mathcal{S} \times \mathcal{A}}
  p^\pi(s^{\prime}, a^{\prime} | s, a)
  \int_{\mathbb{R}}
  F_{\eta (s^{\prime}, a^{\prime})} \left(
    \frac{z-r}{\gamma}
  \right)
  {\rm d}F_{\mathcal{R}(s,a)}(r)
  \label{eq:distributional_Bellman_operator}
  ,
\end{align}
where $
  p^\pi(s^{\prime}, a^{\prime} | s, a) = \mathcal{P}(s^{\prime} | s,a) \pi(a^{\prime} | s^{\prime})
$.
We consider to represent the return distribution by $M\in\mathbb{N}$ atoms.
Let $Z_\theta: S \times A \times \{1,..,M\} \to R$ be a predictor with parameter $\theta$.
In the simplest tabular case, we have $Z_\theta(s,a,m) = \theta(s,a,m)$.
When we estimate the return distribution in the form
$
  \eta_{\theta}(s,a)
  = \frac{1}{M}\sum_{m=1}^{M} \delta_{Z_{\theta}(s,a,m)}
  \in \mathscr{P}_{\theta}(\mathbb{R})^{\mathcal{S} \times \mathcal{A}}
  \subset \mathscr{P}(\mathbb{R})^{\mathcal{S} \times \mathcal{A}}
$,
its corresponding CDF is given by
\begin{align}
  F_{\eta_{\theta}(s,a)}(z)
  =
  \frac{1}{M}\sum_{m=1}^{M}
  \mathbbm{1}\left\{Z_{\theta}(s,a,m) \leq z\right\}
  \label{eq:empirical_return_distribution}
  .
\end{align}
It is convenient if the $m$-th estimator $Z_{\theta}(s,a,m)$ represents the $\tau_m$-th quantile.
The quantile projection operator
$
  \Pi_{w_{1}}:
  \mathscr{P}(\mathbb{R})^{\mathcal{S} \times \mathcal{A}}
  \to
  \mathscr{P}_{\theta}(\mathbb{R})^{\mathcal{S} \times \mathcal{A}}
$
is defined by
\begin{align*}
  \Pi_{w_{1}}(\eta)
  = \arg\min_{\eta_{\theta}\in\mathscr{P}_{\theta}(\mathbb{R})}
  w_{1}(\eta, \eta_{\theta})
  ,
\end{align*}
where
$w_{p}: \mathscr{P}(\mathbb{R}) \to \mathscr{P}(\mathbb{R}) \to [0, \infty]$
is the $p$-th Wasserstein distance defined by
$ 
  w_{p}(\mu, \nu)
  =
  \left(
    \int_{0}^{1} \left|
      F^{-1}_{\mu}(\tau) - F^{-1}_{\nu}(\tau)
    \right|^{p} {\rm d} \tau
  \right)^{1/p}
$ 
for $p\in[1,\infty)$ and thus
$ 
  w_{1}(\eta, \eta_{\theta})
  =
  \sum_{m=1}^{M} \int_{\tau_{m-1}}^{\tau_{m}}
  \left|
    F_{\eta(s,a)}^{-1}(\omega)
    -
    Z_{\theta}(s,a,m)
  \right|
  {\rm d}\omega
$, 
and its minimizers are given as:
$
  Z_{\theta}(s,a,m) = F_{\eta}^{-1}(\hat{\tau}_{m})
$, $
  \hat{\tau}_{m} = \frac{\tau_{m-1}+\tau_{m}}{2}
$ \cite{Dabney2018QR}.
The combined operator $\Pi_{w_{1}} \mathcal{T}^{\pi}$ is $\gamma$-contraction in the supremum $\infty$-Wesserstein distance $\bar{w}_{\infty}$,
where $
  \bar{w}_{p}(\eta, \eta^{\prime})
  =
  \sup_{(s,a)\in \mathcal{S} \times \mathcal{A}}
w_{p}(\eta(s,a), \eta^{\prime}(s,a))
$.
In practice, the quantile projection is implemented by the quantile regression.

\subsection{Offline Reinforcement Learning}

We focuses on offline RL,
where the agent cannot interact with the MDP---instead a batch dataset
$\mathcal{D} = \left\{(s_{i}, a_{i}, r_{i}, s^{\prime}_{i})\right\}_{i=1}^{N}$
of size $|\mathcal{D}| = N$ is given,
where $r \sim \mathcal{R}(s, a)$ and $s^{\prime} \sim \mathcal{P}(\cdot \mid s, a)$.
Let $N(s,a)$ denote the counts that a pair $(s,a)$ appears in $\mathcal{D}$
and $\mathcal{D}(s, a)$ denote the set of all $(r, s^{\prime})$
such that $(s, a, r, s^{\prime}) \in \mathcal{D}$.
For brevity, we assume that $\mathcal{D}$ consists of independent and identically distributed samples.
We denote the empirical Bellman operator by $\mathcal{T}^{\pi}_{\mathcal{D}}$.

In offline RL literature, the principle of \emph{pessimism} is widely used \cite{Buckman2021,Jin2021}, where the value is conservatively estimated so that OOD actions are less chosen
when the learned policy is deployed to the environment.
CODAC \cite{Ma2021} implements the pessimism in offline DRL
by optimizing the following loss:
\begin{align}
  \eta_{k+1}
  = \arg \min_{\eta} \left\{
    \alpha \mathbb{E}_{\tau\sim {\rm Unif}([0,1]), (s,a)\sim\mathcal{D}} \left[
      c_{0}(s,a)F_{\eta}^{-1}(\tau)
    \right]
    + \mathcal{L}_{p}(\eta,\mathcal{T}^{\pi}_{\mathcal{D}}\eta_{k})
  \right\}
  ,
  \label{eq:CODAC_loss}
\end{align}
where $\alpha>0$,
$c_{0}$ is a state-action dependent penalty,
and
$
\mathcal{L}_{p}(\eta,\eta^{\prime})
=
\mathbb{E}_{\mathcal{D}} \left[
  w_{p}(\eta, \eta^{\prime})^{p}
\right]
$.
By the loss \eqref{eq:CODAC_loss},
quantiles are \emph{uniformly} underestimated:
\begin{align}
  F_{\eta_{k+1}(s,a)}^{-1}(\tau)
  =
  F_{\mathcal{T}^{\pi}_{\mathcal{D}}\eta_{k}}^{-1}(\tau)
  - c(s,a)
  ,
  \label{eq:quantiles_shift_by_cde}
\end{align}
where
$c(s,a)$ is a state-action dependent constant, and does \emph{not} depend on $\tau$.

\section{Distorted Distributional Evaluation}

\subsection{Analysis of Offline Distributional Policy Evaluation}

First, we analyze the offline distributional policy evaluation
and motivate the \emph{non-uniform} pessimism.
The missing proofs will be found in Appendix~\ref{ss:appendix:proofs}.
Substituting \eqref{eq:empirical_return_distribution} into the distributional Bellman operator in \eqref{eq:distributional_Bellman_operator},
it holds that
\begin{align*}
  F_{\mathcal{T}^{\pi} \eta_{\theta}(s,a)}(z)
  &=
  \sum_{s^{\prime}, a^{\prime}}
  p^\pi(s^{\prime}, a^{\prime} | s, a)
  \int_{\mathbb{R}}
    F_{\eta_{\theta}(s^{\prime}, a^{\prime})} \left(
      \frac{z-r}{\gamma}
    \right)
  {\rm d}F_{\mathcal{R}(s,a)}(r)
  \\
  &=
  \sum_{s^{\prime}, a^{\prime}}
  p^\pi(s^{\prime}, a^{\prime} | s, a)
  \int_{\mathbb{R}}
    \frac{1}{M}\sum_{m=1}^{M}
    \mathbbm{1}\left\{
      Z_{\theta}(s^{\prime}, a^{\prime}, m) \leq \frac{z-r}{\gamma}
    \right\}
  {\rm d}F_{\mathcal{R}(s,a)}(r)
  \\
  &=
  \underset{\substack{
      R \sim \mathcal{P}(\cdot | s,a), \\
      S^{\prime} \sim \mathcal{R}(\cdot | s,a), \\
      A^{\prime} \sim \pi(\cdot | S^{\prime})
  }}{
    \mathbb{E}
  }
  \left[
    \frac{1}{M}\sum_{m=1}^{M}
    \mathbbm{1}\left\{
      R + \gamma Z_{\theta}(S^{\prime}, A^{\prime}, m) \leq z
    \right\}
  \right]
  .
\end{align*}
In the above expressions,
both the CDF $F_{\eta_{\theta}(s^{\prime}, a^{\prime})} \left(\frac{z-r}{\gamma}\right)$
and the PDF ${\rm d}F_{\mathcal{R}(s,a)}(r)$ are (locally) integrable.
In addition, given that the reward function is bounded, they both have bounded support.
Therefore, the convolution above is well-defined and
$F_{\mathcal{T}^{\pi} \eta_{\theta}(s,a)}(z)$ is continuous.

In offline RL, we approximate the expectation using the dataset $\mathcal{D}$ by
\begin{align}
  &F_{\mathcal{T}^{\pi}_{\mathcal{D}} \eta_{\theta}(s,a)}(z)
  \nonumber
  \\
  &=
  \underset{\substack{
      (S^{\prime},R)\sim\mathcal{D}(s,a), \\
      A^{\prime} \sim \pi(\cdot | S^{\prime})
  }}{
    \mathbb{E}
  }
  \left[
    \frac{1}{M}\sum_{m=1}^{M}
    \mathbbm{1}\left\{
      R + \gamma Z_{\theta}(S^{\prime}, A^{\prime}, m) \leq z
    \right\}
  \right]
  \nonumber
  \\
  &=
  \frac{1}{N(s,a)}
  \sum_{(s^{\prime},r)\in\mathcal{D}(s,a)}
  \underbrace{
    \sum_{a^{\prime}\in\mathcal{A}} \pi(a^{\prime}|s^{\prime})
      \frac{1}{M} \sum_{m=1}^{M}
      \mathbbm{1} \left\{
        r + \gamma Z_{\theta}(s^{\prime},a^{\prime},m) \leq z
      \right\}
    }_{
    \tilde{F}_{\theta,\pi}(z|r, s^{\prime})
  }
  \label{eq:empricial_Bellman_target}
  .
\end{align}

Given the expression above,
we investigate the effect of one-step distributional evaluation.
We assume the following properties on $F_{\mathcal{T}^{\pi} \eta_{\theta}(s,a)}$.
\begin{assumption}
  \label{assumption:bellman_target_is_smooth}
  $F_{\mathcal{T}^{\pi} \eta_{\theta}(s,a)}(z)$ is smooth
  for all $(s,a)\in\mathcal{S}\times\mathcal{A}$ and $\theta$.
\end{assumption}
\begin{assumption}
  \label{assumption:bellman_target_is_strictly_increasing}
  $\frac{\rm d}{{\rm d} z} F_{\mathcal{T}^{\pi} \eta_{\theta}(s,a)}(z) > 0$
  for all $(s,a)\in\mathcal{S}\times\mathcal{A}$ and $\theta$.
\end{assumption}

Let $f_{\pi,\theta}$ be the density of $F_{\mathcal{T}^{\pi} \eta_{\theta}(s,a)}$.
The following theorem warns that the error decay is not uniform in $\tau$ even in the asymptotic case $N\to\infty$.
\begin{restatable}[Asymptotic Behavior of Distributional Evaluation]{theorem}{CLTofEmpiricalBellmanTarget}
  \label{claim:clt_of_empirical_bellman_target}
  Let
  Assumptions \ref{assumption:bellman_target_is_smooth} and \ref{assumption:bellman_target_is_strictly_increasing}
  hold.
  For the estimator \eqref{eq:empricial_Bellman_target},
  we have
  \begin{align*}
    \sqrt{N(s,a)} \left(
      F_{\mathcal{T}^{\pi}_{\mathcal{D}} \eta_{\theta}(s,a)}^{-1}(\tau)
      -
      F_{\mathcal{T}^{\pi} \eta_{\theta}(s,a)}^{-1}(\tau)
    \right)
    \to
    \mathcal{N}(0, \tilde{\sigma}_{\tau}^{2}(s,a))
  \end{align*}
  as $N(s,a)\to\infty$,
  where we define $\tilde{\sigma}_{\tau}^{2}(s,a)$ using the empirical CDF $\tilde F$ in \eqref{eq:empricial_Bellman_target} as
  \begin{align*}
    \tilde{\sigma}_{\tau}^{2}(s,a)
    =
    \frac{
      \mathbb{V}\left[\tilde{F}_{\theta,\pi}(F_{\mathcal{T}^{\pi} \eta_{\theta}(s,a)}^{-1}(\tau)|R, S^{\prime})|s,a\right]
    }{
      f_{\pi,\theta}(z_\tau)^2
    }.
  \end{align*}
  Moreover, we have
  \begin{align*}
    \frac{\varepsilon}{f_{\pi,\theta}(z_\tau)^2}
    \leq
    \tilde{\sigma}_{\tau}^{2}(s,a)
    \leq
    \frac{
        \tau(1-\tau)
    }{
      f_{\pi,\theta}(z_\tau)^2
    }
  \end{align*}
  for a constant $\varepsilon\ge 0$, and $\varepsilon>0$ holds when
  $\mathbb{P}(R\ge z_\tau - \gamma \overline{z})$ and $\mathbb{P}(R< z_\tau - \gamma \underline{z})$ are nonzero for
  \begin{align}
    \overline{z} & \coloneqq \max_{s\in\mathcal{S},a\in\mathcal{A},m\le \frac{2}{3}M} Z_\theta(s,a,m) \label{eq:defL} \\
    \underline{z} & \coloneqq \min_{s\in\mathcal{S},a\in\mathcal{A},m> \frac{M}{3}} Z_\theta(s,a,m) \label{eq:defU}.
\end{align}
\prooflink{clt_of_empirical_bellman_target}
\end{restatable}

Next, we provide a point-wise concentration bound of
$F_{\mathcal{T}^{\pi}_{\mathcal{D}} \eta_{\theta}(s,a)}(z)$.
\begin{theorem}[Point-wise Concentration of Distributional Evaluation]
  \label{claim:density_aware_concentration_of_ede}
  Let Assumptions \ref{assumption:bellman_target_is_smooth}
  and \ref{assumption:bellman_target_is_strictly_increasing}
  hold,
  $
    \Delta(s,a,z)
    = \frac{1}{f_{\pi,\theta}(z)}
    \sqrt{
      \coeffC
    }
  $
  and
  $z_{\tau} = F_{\mathcal{T}^{\pi} \eta_{\theta}(s,a)}^{-1}(\tau)$.
  Then,
  for $\delta > 0$ and for all $(s,a)\in\mathcal{S}\times\mathcal{A}$,
  with probability at least $1-\delta$, we have
  \begin{align}
    \left|
    F_{\mathcal{T}^{\pi} \eta_{\theta}(s,a)}^{-1}(\tau)
    -
    F_{\mathcal{T}^{\pi}_{\mathcal{D}} \eta_{\theta}(s,a)}^{-1}(\tau)
    \right|
    <
    \Delta(s,a,\bar{z}_{\tau})
    \label{eq:pointwise_concentration_of_DDE_vague}
    ,
  \end{align}
  where $\bar{z}_{\tau}$ is a $\Delta(s,a,z_\tau)$-neighborhood
  of $z_\tau$.
  In addition, if $f_{\pi,\theta}$ is $\alpha$-Lipschits for some $\alpha\ge 0$
  and $
    N(s,a) \ge \frac{2\alpha^2}{f(z_\tau)^4} \log\frac{2|\mathcal{S}||\mathcal{A}|}{\delta}
  $,
  we have
  \begin{align}
    \left|
    F_{\mathcal{T}^{\pi} \eta_{\theta}(s,a)}^{-1}(\tau)
    -
    F_{\mathcal{T}^{\pi}_{\mathcal{D}} \eta_{\theta}(s,a)}^{-1}(\tau)
    \right|
    <
    2 \Delta(s,a,z_{\tau})
    \label{eq:pointwise_concentration_of_DDE_specific}
    .
  \end{align}
  \prooflink{density_aware_concentration_of_ede}
\end{theorem}
The bounds
\eqref{eq:pointwise_concentration_of_DDE_vague} and
\eqref{eq:pointwise_concentration_of_DDE_specific} suggests that
the confidence bound is governed by not only the sample size $N$
but also the inverse density $1/f_{\pi,\theta}$.
When $f_{\pi,\theta}$ is close to $0$,
the bounds get extremely large.
This must be the case at distributional tails, where $\tau$ close to $0$ or $1$.
Therefore, uniform pessimism would result in over conservatism,
which hurts generalization and performance in practice.

\subsection{Distortion Operator}

Let $\mathcal{G} \subseteq \mathbb{R}$ be a set and
$
  \phi:
  \mathcal{S} \times \mathcal{A} \times [0,1] \to \mathcal{G}
$
be a function, both of which we specify later.
We define a {\itshape quantile distortion operator}
$
  \mathcal{Q}_{\phi}:
  \mathscr{P}(\mathbb{R})^{\mathcal{S} \times \mathcal{A}}
  \to
  \mathscr{P}(\mathbb{R})^{\mathcal{S} \times \mathcal{A}}
$
associated with a function $\phi$ by
\begin{align}
  F_{\mathcal{Q}_{\phi}\eta(s, a)}^{-1}(\tau)
  =
  F_{\eta(s, a)}^{-1}(\tau) - \phi(s,a,\tau)
  .
  \label{eq:LCB:quantile_distortion}
\end{align}
Then the distribution $\mathcal{Q}_{\phi} \eta_{\theta}$ can be expressed as
\begin{align}
  \mathcal{Q}_{\phi} \eta_{\theta}(s,a)
  =
  \frac{1}{M}\sum_{m=1}^{M} \delta_{Z_{\theta}(s,a,m) - \phi(s,a,\tau_{m})}
  .
\end{align}
The operator $\mathcal{Q}_{\phi}$ not only shifts but also distorts the quantile function.
Indeed, unlike the property of CODAC shown in \eqref{eq:quantiles_shift_by_cde},
the transformation in \eqref{eq:LCB:quantile_distortion} is \emph{not uniform}
in the sense that it depends on $\tau$ through $\phi$.
If we choose non-negative $\phi\in\mathcal{G}=\mathbb{R}_{\ge 0}$, the operator $\mathcal{Q}_{\phi}$ pushes down the quantile function of $\eta$
and therefore gives a \emph{non-uniform} pessimistic estimate of $\eta$ depending on $\phi$.

It is easily shown that
the distortion operator $\mathcal{Q}_{\phi}$ is non-expansion.
Indeed, for $\mu,\nu\in\mathscr{P}(\mathbb{R})^{\mathcal{S} \times \mathcal{A}}$,
it holds that
\begin{align*}
  F_{\mathcal{Q}_{\phi}\mu(s, a)}^{-1}(\tau) -
  F_{\mathcal{Q}_{\phi}\nu(s, a)}^{-1}(\tau)
  &=
  F_{\mu(s, a)}^{-1}(\tau) - \phi(s,a,\tau) -
  F_{\nu(s, a)}^{-1}(\tau) + \phi(s,a,\tau)
  \\
  &=
  F_{\mu(s, a)}^{-1}(\tau) -
  F_{\nu(s, a)}^{-1}(\tau)
  .
\end{align*}
Since $\mathcal{T}^{\pi}$ is a $\gamma$-contraction in $\bar{w}_{p}$,
the following proposition holds immediately.
\begin{proposition}[Contraction]
  \label{pp:dde_is_contraction}
  The composition $\mathcal{Q}_{\phi} \mathcal{T}^{\pi}$ is
  a $\gamma$-contraction in $\bar{w}_{p}$.
  Furthermore,
  the combined operator $\Pi_{w_{1}} \mathcal{Q}_{\phi} \mathcal{T}^{\pi}$
  is a $\gamma$-contraction in $\bar{w}_{\infty}$.
\end{proposition}

Now, we define {\itshape distorted distributional evaluation} (DDE) and its quantile-projected variant by
\begin{align*}
  \eta_{{k+1}}
  &=
  \mathcal{Q}_{\phi} \mathcal{T}^{\pi} \eta_{{k}}
  \quad\text{and}\quad
  \eta_{\theta_{k+1}}
  =
  \Pi_{w_{1}} \mathcal{Q}_{\phi} \mathcal{T}^{\pi} \eta_{\theta_{k}}
  ,
\end{align*}
respectively.
The following theorem characterizes the fixed point of DDE.
\begin{theorem}
  [Fixed Point of Distorted Distributional Evaluation]
  \label{claim:fixed_point_of_dde}
  Let $\overline \phi \equiv \sup_{s,a,\tau} \phi(s,a,\tau)$
  and $\underline{\phi} \equiv \inf_{s,a,\tau} \phi(s,a,\tau)$.
  Let $F_\infty^{-1}$ and $\underline F_\infty^{-1}$ be the fixed points of $\mathcal{T}^{\pi}$ and $\mathcal{Q}_{\phi} \mathcal{T}^{\pi}$, respectively.
  Then we have
  \begin{align*}
    F^{-1}_\infty(\tau;s,a) - \phi(s,a,\tau) - \frac{\gamma}{1-\gamma} \, \overline\phi
    &\le \underline F_\infty^{-1}(\tau;s,a)
    \\
    &\le F^{-1}_\infty(\tau;s,a) - \phi(s,a,\tau) - \frac{\gamma}{1-\gamma} \, \underline\phi
    .
  \end{align*}
  \prooflink{fixed_point_of_dde}
\end{theorem}

\subsection{Quantile-wise Pessimism via Distortion}

\subsubsection{Evaluation.}
Combining Eq. \eqref{eq:LCB:quantile_distortion}
and Theorem \ref{claim:density_aware_concentration_of_ede}, we have,
w.p. $1-\delta$,
\begin{align*}
  F_{\mathcal{T}^{\pi} \eta_{\theta}(s,a)}^{-1}(\tau)
  >
  F_{\mathcal{Q}_{\phi} \mathcal{T}^{\pi}_{\mathcal{D}} \eta_{\theta}(s,a)}^{-1}(\tau)
  +
  \phi(s,a,\tau) -
  \frac{1}{f_{\pi,\theta}(\bar{z}_{\tau})}
  \sqrt{
    \coeffC
  }
  .
\end{align*}
The above expression suggests that,
if we choose a function $\phi$ such that
\begin{align}
  \phi(s,a,\tau)
  \ge
  \frac{1}{f_{\pi,\theta}(\bar{z}_{\tau})}
  \sqrt{\coeffC}
  \label{eq:requirement_for_phi}
  ,
\end{align}
then we obtain a point-wise pessimistic estimate of quantile function from offline data.
This is advantageous because the information of distribution tail might less appear in the offline dataset.
A challenge in \eqref{eq:requirement_for_phi}
even for tabular settings is that,
we need the density ${f_{\pi,\theta}}$ of the unknown distribution $F_{\mathcal{T}^{\pi} \eta_{\theta}(s,a)}(z)$.

We propose to
(i) use an ensemble of $L\in\mathbb{N}$ predictors for distributional evaluation,
where
each predictor $Z_{\theta_{\ell}}$ with parameter $\theta_{\ell}$, $\ell\in\{1,\ldots,L\}$,
outputs the quantile values for
$\ell$-th estimate of the return distribution $\eta_{\ell}$ as
$F_{\eta_{\ell}(s,a)}^{-1}(\tau_{m}) = Z_{\theta_{\ell}}(s,a,m)$,
and
(ii) use the standard deviation of $F_{\eta_{\ell}(s,a)}^{-1}(\tau_{m})$
over $L$ predictors, $\hat{\sigma}(s,a,{\tau_m})$,
as a surrogate
for \eqref{eq:requirement_for_phi} for the point $\tau_{m}$.
To be precise,
we choose
\begin{align*}
  \phi(s,a,\tau_m)
  =
  \beta \hat{\sigma}(s,a,{\tau_m})
  ,
\end{align*}
where $\beta > 0$ is a hyperparameter.
At a point $(s,a,\tau_m)$ where $f_{\pi,\theta}(z_{\tau_m})$ and $N(s,a)$ are \emph{large},
$z_{\tau_m} = F_{\mathcal{T}^{\pi} \eta_{\theta}(s,a)}^{-1}(\tau_m)$
would be \emph{more} predictable and $\hat{\sigma}(s,a,{\tau_m})$ would get \emph{small} value.
On the other hand,
at a point where $f_{\pi,\theta}(z_{\tau_m})$ and $N(s,a)$ are \emph{small},
$z_{\tau_m}$
would be \emph{less} predictable and $\hat{\sigma}(s,a,{\tau_m})$ would get \emph{large} value.
Though \eqref{eq:requirement_for_phi} requires $\phi$ to be chosen depending on the current distributions $\eta_{\theta_\ell}(s,a)$,
we hypothesize that a fast-varying $\phi$ may hurt the learning stability in practice.
Thus, we use target networks with parameter $\{\bar{\theta}_{\ell}\}$ that are updated slowly, i.e.,
$\bar{\theta_{\ell}} \leftarrow (1-\kappa) \bar{\theta_{\ell}} + \kappa \theta_{\ell}$ with $\kappa\in(0,1)$, to compute $\hat{\sigma}(s,a,{\tau_m})$.
The updated of ensemble $\{\theta_{\ell}\}$ is conducted as follows.
Let $\rho^{\kappa}_{\tau}$ be the quantile Huber loss defined by
\begin{align*}
  \rho_{\tau}^{\kappa}(u)
  &=
  \left| \tau - \mathbbm{1}{\{u<0\}} \right| \mathcal{L}_{\kappa}(u)
  ,
  \quad\text{where}\quad
  \mathcal{L}_{\kappa}(u)
  &= \left\{
  \begin{array}{ll}
  \frac{1}{2}u^2 & \text{if} \, |u| < \kappa \\
  \kappa \left(|u|-\frac{1}{2}\kappa\right) & \text{otherwise}
  \end{array}
  \right.
  .
\end{align*}
The $\ell$-th predictor is updated by minimizing the following loss:
\begin{align*}
  &\mathcal{L}(\theta_{\ell})
  =
  \mathbb{E}_{(s,a,s^{\prime},r)\sim\mathcal{D}}\left[
  \frac{1}{M}
  \sum_{1 \leq i,j \leq M}
  \rho^{\kappa}_{\hat{\tau}_{i}}
  \bigl(
    \mathcal{T} Z(s,a,s^{\prime},r,j) - Z_{\theta_{\ell}}(s,a,i)
  \bigr)
  \right]
  ,
  \nonumber
  \\
  &\mathcal{T} Z(s,a,s^{\prime},r,j)
  =
  r + \gamma \hat{\mu}(s^{\prime},a^{\prime},j)
  - \beta \hat{\sigma}(s,a,{j})
  , \quad
  a^{\prime} \sim \pi(\cdot|s^{\prime})
  ,
\end{align*}
where
$\hat{\mu}(s,a,j)$ and $\hat{\sigma}(s,a,{j})$
are the empirical mean and the empirical standard deviation
of $\tau_{j}$-th quantile value $\left\{ Z_{\bar{\theta}_{\ell}}(s,a,j) \right\}_{\ell=1}^{L}$ outputted by the target networks, respectively,
and
$\hat{\tau}_{i}=\frac{\tau_{i-1}+\tau_{i}}{2}$.

\vspace{-10pt}
\subsubsection{Control.}
For discrete action spaces, we approximate the action value as $
  Q = \frac{1}{LM} \sum_{\ell=1}^{L} \sum_{m=1}^{M}
    Z_{\theta_{\ell}}(s,a,m)
$ and let $\pi$ be $\epsilon$-greedy, for example.
For continuous action spaces,
we adopt a stochastic actor $\pi_{\psi}$ with tractable distribution and update its parameter $\psi$ by a loss
$
  J(\psi) = \mathbb{E}_{s\sim\mathcal{D}, a\sim\pi_{\psi}(\cdot|s)} \left[
    \alpha\log\pi_{\psi}(a|s) - Q(s,a)
  \right]
$,
where $\alpha>0$.
For both cases, it is easy to extend to risk-sensitive settings if we replace $Q$ with the distorted expectation as in \cite{Dabney2018IQ}.
We name an actor critic method based on DDE as Distorted Distributional Actor Critic (DDAC).

\section{Experiment}

\subsubsection{Setup.}

We compare the proposed method and CODAC on \texttt{InvManagement-v1} of OR-Gym \cite{HubbsOR-Gym}.
To prepare an offline dataset,
we first trained an expert by PPO \cite{Schulman2017} and then sampled its trajectories and mixed them with trajectories of a random policy.
Both of CODAC and DDAC are built on top of SAC \cite{Haarnoja2019} with quantile-based distributional critic \cite{Dabney2018QR} with $M=32$ atoms.
Critic and actor are represented by fully-connected feed-forward networks with two hidden layers and each hidden layer has $256$ units with ReLU activations.
The hyperhapameters are set to the default values of CODAC and SAC unless specified, respectively.
For DDAC, we used $L=10$ predictors for ensemble.
We conducted a sweep over $\beta\in\{0.1,0.3,0.5,0.7,0.9\}$
and found that $\beta=0.5$ works best.
CODAC has two hyperparameters $\omega$ and $\zeta$, which are used to adjust the degree of pessimism $\alpha$ in the loss \eqref{eq:CODAC_loss} adaptively.
We conducted a sweep over $\omega\in\{0.1,1,10\}$ and $\zeta\in\{1,10\}$
and found that $\omega=1$ and $\zeta=10$ works best.
The independent learning trials are conducted with $10$ different ranodm seeds.

\subsubsection{Results.}

\begin{figure}
  \includegraphics[width=\textwidth]{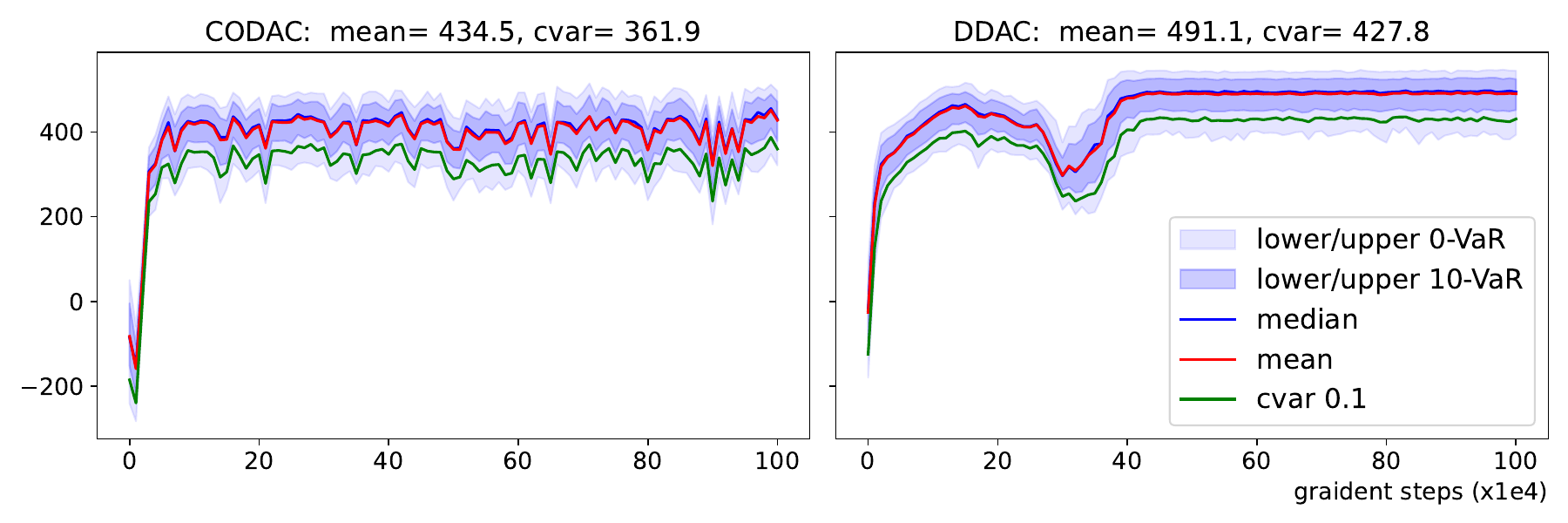}
  \vspace{-20pt}
  \caption{Learning result in \texttt{InvManagement-v1}.}
  \label{fig:inv_LC}
  \vspace{-20pt}
\end{figure}

Figure \ref{fig:inv_LC} compares the learning result of CODAC and DDAC for best hyperparameter values.
The distributions of test performances in the environment 
averaged over 10 seeds are plotted over the gradient steps.
DDAC learnes stabler than CODAC, and the final performance was also better both in terms of mean and CVaR(0.1).

\vspace{-10pt}

\section{Conclusion}

In this work, we statistically analyzed offline distributional evaluation and proposed DDE, that enables non-uniform pessimism through the notion of quantile distortion.
We evaluated our actor critic method, DDAC, in a simulated benchmark and showed that DDAC learns stabler and reaches better final performance than the existing uniform pessimism method, CODAC.

Our proposed method is easily applied to risk-sensitive settings.
Since it is straightforward to compute various risk measures from the estimated distorted distribution, our method would work particularly well in search based AutoRL method \cite{Marinescu2022}.
We believe that this paper paves the way for real-world applications of offline DRL.

%
%
\bibliographystyle{splncs04}
\bibliography{references}

\newpage

\appendix
\section{Proofs}
\label{ss:appendix:proofs}

\subsection{Proof of Theorem \ref{claim:clt_of_empirical_bellman_target}}
\begin{proof}
\label{proof:clt_of_empirical_bellman_target}
For brevity, fix $s,a,\pi,\theta$ and let
$F\coloneqq F_{\mathcal{T}^\pi\eta_\theta(s,a)}$,
$\hat F\coloneqq F_{\mathcal{T}^{\pi}_{\mathcal{D}} \eta_{\theta}(s,a)}$,
$f=f_{\pi,\theta}$,
$z_\tau\coloneqq F^{-1}(\tau)$,
$\hat z_\tau\coloneqq \hat F^{-1}(\tau)$, and
$N\coloneqq |\mathcal{D}(s,a)|$.  Recall that our estimator is the average of $N$ i.i.d.\ samples,
  \begin{align*}
    \hat F(z)
    &=
    \frac{1}{N}
    \sum_{(s^{\prime},r)\in\mathcal{D}(s,a)}
    \underbrace{
      \sum_{a^{\prime}\in\mathcal{A}} \pi(a^{\prime}|s^{\prime})
        \frac{1}{M} \sum_{m=1}^{M}
        \mathbbm{1} \left\{
          r + \gamma Z_{\theta}(s^{\prime},a^{\prime},m) \leq z
        \right\}
      }_{
      \tilde{F}_{\theta,\pi}(z|r, s^{\prime})
    },
  \end{align*}
and that the corresponding expectation is $F(z)=\E[\tilde{F}_{\theta,\pi}(z|R, S^{\prime})]$, where the expectation is with respect to the random next state $S^{\prime}$ and random reward $R$ given that action $a$ is taken at state $s$.

Since $f$ is continuous, there exists $\bar z_\tau\in[\hat z_\tau \land z_\tau, \hat z_\tau \lor z_\tau,]$ such that
\begin{align}
    F(\hat z_\tau) = F(z_\tau) + f(\bar z_\tau) (\hat z_\tau - z_\tau)
    \label{eq:mean-value}
\end{align}
by the mean value theorem. Hence,
\begin{align*}
    \hat F^{-1}(\tau) - F^{-1}(\tau)
    & = \hat z_\tau - z_\tau \\
    & = \frac{1}{f(\bar z_\tau)} (F(\hat z_\tau) - F(z_\tau))
        &
        \mbox{(by \eqref{eq:mean-value})} \\
    & = \frac{1}{f(\bar z_\tau)} (F(\hat z_\tau) - \hat F(\hat z_\tau))
    + \frac{1}{f(\bar z_\tau)} (\hat F(\hat z_\tau) - F(z_\tau))\\
    & = \frac{1}{f(\bar z_\tau)} (F(\hat z_\tau) - \hat F(\hat z_\tau))
    + \frac{1}{f(\bar z_\tau)} (\hat F(\hat z_\tau) - \tau)
    ,
\end{align*}
where the last equality holds from the continuity of $F$.
We use this expression to study the limit of $\sqrt{N}(\hat F^{-1}(\tau) - F^{-1}(\tau))$ as $N\to\infty$.
Since $\hat F$ is an empirical distribution of $N$ continuous random variables, we have $|\hat F(\hat z_\tau) - \tau| \le 1/N$ almost surely.
Hence,
$ 
    \sqrt{N} (\hat F(\hat z_\tau) - \tau)
    \to 0
$ 
as $N\to\infty$.  Since $\hat z_\tau \to z_\tau$ as $N\to\infty$ almost surely by the strong law of large numbers, the continuity of $f$ and the central limit theorem imply
\begin{align*}
    \sqrt{N}\frac{F(\hat z_\tau) - \hat F(\hat z_\tau)}{f(\bar z_\tau)}
    \to \mathcal{N}\left(0,\frac{\tilde\sigma^2(z_\tau)}{f(z_\tau)^2}\right)
\end{align*}
where
$
  \tilde{\sigma}^{2}(z)
  =
  \mathbb{V}\left[\tilde{F}_{\theta,\pi}(z|R, S^{\prime})\right]
$.
These imply, as desired,
\begin{align*}
  \sqrt{N}(\hat F^{-1}(\tau) - F^{-1}(\tau))
  \to \mathcal{N}\left(0,\frac{\tilde\sigma^2(z_\tau)}{f(z_\tau)^2}\right)
  .
\end{align*}

Now, we proceed to bounding $\tilde{\sigma}^{2}(F^{-1}(\tau))$.
For an upper bound, we use Bhatia-Davis inequality, which states that,
for a distribution which has a bounded support $[m, M]$ and a mean $\mu$,
the variance $\sigma^{2}$ is upper-bounded by
$
  \sigma^{2} \leq (M-\mu)(\mu-m)
$.
Then, observing that
$\tilde{F}_{\theta,\pi}(z|r, s^{\prime}) \in [0,1]$
and
$
  \mathbb{E} \left[
    \tilde{F}_{\theta,\pi}(z|R, S^{\prime})
  \right]
  =
  F(z)
  $,
we have
$ 
  \tilde{\sigma}^{2}(z)
  =
  \mathbb{V}\left[\tilde{F}_{\theta,\pi}(z|R, S^{\prime})\right]
  \leq
  \left(1 - F(z)\right)
            F(z)
  .
$ 
Letting $z=z_{\tau}=F^{-1}(\tau)$,
we obtain
\begin{align*}
  \tilde{\sigma}^{2}\left(F^{-1}(\tau)\right)
  \leq
  \left(1 - F\left(F^{-1}(\tau)\right)\right)
            F\left(F^{-1}(\tau)\right)
  = \tau(1 - \tau)
  .
\end{align*}
A lower bound with $\varepsilon=0$ trivially holds.  Hence, in the following, we show $\tilde{\sigma}^{2}\left(F^{-1}(\tau)\right)>\varepsilon$ for an $\varepsilon>0$ under the condition given in the theorem.
By \eqref{eq:defL}--\eqref{eq:defU}, we have
\begin{align*}
  \tilde F_{\theta,\pi}(z|r,s')
  & \ge \sum_{a',m\le 2M/3} \frac{\pi(a'|s')}{M} \mathbbm{I}\left\{r + \gamma Z_\theta(s',a',m) \le z\right\}\\
  & \ge \sum_{a',m\le 2M/3} \frac{\pi(a'|s')}{M} \mathbbm{I}\left\{r + \gamma \overline{z} \le z\right\}
  & \mbox{by \eqref{eq:defL}} \\
  & \ge \frac{2}{3} \mathbbm{I}\{r+\gamma \overline{z}\ge z\},
\end{align*}
and
\begin{align*}
  \tilde F_{\theta,\pi}(z|r,s')
  & \le \frac{1}{3} + \sum_{a',m> M/3} \frac{\pi(a'|s')}{M} \mathbbm{I}\left\{r + \gamma Z_\theta(s',a',m) \le z\right\}\\
  & \le \frac{1}{3} + \sum_{a',m> M/3} \frac{\pi(a'|s')}{M} \mathbbm{I}\left\{r + \gamma \underline{z} \le z\right\}
  & \mbox{by \eqref{eq:defU}} \\
  & \le \frac{1}{3} + \frac{1}{3} \mathbbm{I}\{r+\gamma \underline{z}\ge z\}.
\end{align*}
Hence,
$\tilde F_{\theta,\pi}(z|R,S') \ge 2/3$ with probability at least $\mathbb{P}(R\ge z - \gamma \overline{z})$, and
$\tilde F_{\theta,\pi}(z|R,S') \le 1/3$ with probability at least $\mathbb{P}(R< z - \gamma \underline{z})$.
Letting $\mu\coloneqq\E[\tilde F_{\theta,\pi}(z|R,S')]$ and $\varepsilon'=\min\{\mathbb{P}(R\ge z - \gamma \overline{z}),\mathbb{P}(R< z - \gamma \underline{z})\}$, we have
\begin{align*}
  \mathbb{V}\left[\tilde F_{\theta,\pi}(z|R,S')\right]
  & \ge \left(\frac{2}{3}-\mu\right)_+^2 \mathbb
  {P}(R\ge z - \gamma \overline{z})
  + \left(\mu-\frac{1}{3}\right)_+^2 \mathbb
  {P}(R< z - \gamma \underline{z}) \\
  & \ge \left(\left(\frac{2}{3}-\mu\right)_+^2 + \left(\mu-\frac{1}{3}\right)_+^2\right) \varepsilon' \\
  & \ge \frac{\varepsilon'}{18}.
\end{align*}
Since $\varepsilon'>0$, this establishes the lower bound.
\prooflinkfinal{clt_of_empirical_bellman_target}
\end{proof}

\subsection{Proof of Theorem \ref{claim:density_aware_concentration_of_ede}}
\begin{proof}
\label{proof:density_aware_concentration_of_ede}
  We follow a similar argument as Proposition 2 of \cite{Kolla2019},
  though we focus on obtaining a point-wise bound instead of a uniform bound.
  Let $\Delta: \mathcal{S} \times \mathcal{A} \times \mathbb{R} \to \mathbb{R}_{>0}$.
  For a fixed state-action pair $(s,a)$, let us denote
  $F = F_{\mathcal{T}^{\pi} \eta_{\theta}(s,a)}$,
  $\hat{F} = F_{\mathcal{T}^{\pi}_{\mathcal{D}} \eta_{\theta}(s,a)}$,
  $\epsilon_{\tau} = \Delta(s,a,z_\tau)$, and
  $z_{\tau} = F^{-1}(\tau)$.
  Then, for a fixed $\tau\in[0,1]$, we have
  \begin{align*}
    &\mathbb{P} \left[
      \left|
      F^{-1}(\tau)
      -
      \hat F^{-1}(\tau)
      \right|
      \ge
      \epsilon_\tau
    \right]
    \\
    &=
    \mathbb{P} \left[
      \hat{F}^{-1}(\tau)
      \ge
        z_{\tau} + \epsilon_{\tau}
    \right]
    +
    \mathbb{P} \left[
      \hat{F}^{-1}(\tau)
      \leq
        z_{\tau} - \epsilon_{\tau}
    \right]
    \\
    &=
    \mathbb{P} \left[
      \hat{F}\left(
        \hat{F}^{-1}\left(\tau
      \right)\right)
      \ge
      \hat{F}\left(
        z_{\tau} + \epsilon_{\tau}
      \right)
    \right]
    +
    \mathbb{P} \left[
      \hat{F}\left(
        \hat{F}^{-1}\left(\tau
      \right)\right)
      \leq
      \hat{F}\left(
        z_{\tau} - \epsilon_{\tau}
      \right)
    \right]
    \\
    &\leq
    \mathbb{P} \left[
      \tau
      \ge
      \hat{F}\left(
        z_{\tau} + \epsilon_{\tau}
      \right)
    \right]
    +
    \mathbb{P} \left[
      \tau
      \leq
      \hat{F}\left(
        z_{\tau} - \epsilon_{\tau}
      \right)
    \right]
    \\
    &=
    \mathbb{P} \left[
      F\left(
        z_{\tau} + \epsilon_{\tau}
      \right)
      - \hat{F}\left(
        z_{\tau} + \epsilon_{\tau}
      \right)
      \ge
      F\left(
        z_{\tau} + \epsilon_{\tau}
      \right)
      - \tau
    \right]
    \\
    &\quad+
    \mathbb{P} \left[
      \hat{F}\left(
        z_{\tau} - \epsilon_{\tau}
      \right)
      -
      F\left(
        z_{\tau} - \epsilon_{\tau}
      \right)
      \ge
      \tau
      -
      F\left(
        z_{\tau} - \epsilon_{\tau}
      \right)
    \right]
    .
  \end{align*}
  We proceed to obtain the upper bound for the last expression.
  Since $\tilde{F}(z|r, s^{\prime}) \in [0,1]$,
  applying Hoeffding's inequallity to each tail individually, we obtain
  \begin{align}
    \mathbb{P} \left[
      \left|
      F^{-1}(\tau)
      -
      \hat F^{-1}(\tau)
      \right|
      \ge
      \epsilon_\tau
    \right]
    &\leq
    \exp\left(-2 N(s,a) \left(
      F\left(z_{\tau}\!+\!\epsilon_{\tau}\right) - \tau
    \right)^2\right)
    \nonumber\\
    &\quad
    +
    \exp\left(-2 N(s,a) \left(
      \tau - F\left(z_{\tau}\!-\!\epsilon_{\tau}\right)
    \right)^2\right)
    \label{eq:density_aware_concentration_of_ede:after_hoefffding}
    \\
    &\leq
    2\exp\left(-2 N(s,a) \bar{\epsilon}_{\tau}^2 \right)
    \nonumber
    ,
  \end{align}
  where
  $ 
    \bar{\epsilon}_{\tau}
    = \min \left\{
      F\left(z_{\tau}
       + \epsilon_{\tau}\right) - F\left(z_{\tau}\right)
      ,
      F\left(z_{\tau}\right)
       - F\left(z_{\tau} - \epsilon_{\tau}\right)
    \right\}
  $ 
  with $\tau=F\left(z_{\tau}\right)$.
  Letting $f$ be the density of $F$,
  we have, from the intermediate value theorem, that
  $
    \bar{\epsilon}_{\tau}
    = \epsilon_{\tau} \min \left\{
      f(z'), f(z'')
    \right\}
  $
  for some
  $z' \in [z_{\tau}, z_{\tau} + \epsilon_{\tau}]$ and
  $z'' \in [z_{\tau} - \epsilon_{\tau}, z_{\tau}]$.
  Therefore, by applying union bound,
  with probability at least $1-\delta$, we have
  \begin{align*}
    \left|
    F^{-1}(\tau)
    -
    \hat F^{-1}(\tau)
    \right|
    <
    \Delta(s,a,\bar{z}_{\tau})
  \end{align*}
  for all $(s,a)\in\mathcal{S}\times\mathcal{A}$, where
  $ 
    \Delta(s,a,z)
    =
    \frac{1}{f(z)}
    \sqrt{
      \coeffC
    }
  $ 
  and $\bar{z}_{\tau}$ is an $\epsilon_\tau$-neighborhood of
  $F^{-1}(\tau)$.

  Now, for a constant $\alpha>0$, let
  $ 
      N \ge \frac{2\alpha^2}{f(z_\tau)^4} \log\frac{2|\mathcal{S}||\mathcal{A}|}{\delta}.
  $ 
  Then we have
  \begin{align}
      \epsilon_\tau
      = \frac{1}{f(z_\tau)} \sqrt{\frac{1}{2N} \log\frac{2|\mathcal{S}||\mathcal{A}|}{\delta}}
      \le \frac{f(z_\tau)}{2\alpha}.
      \label{eq:epsilon-UB}
  \end{align}
  Therefore, since $|\bar z_\tau -z_\tau|\le \epsilon_\tau$, we have
  \begin{align*}
      \Delta(s,a,\bar z_\tau)
      & = \frac{1}{f(\bar z_\tau)} \sqrt{\frac{1}{2N} \log\frac{2|\mathcal{S}||\mathcal{A}|}{\delta}}
      \\
      & \le \frac{1}{f(z_\tau)-\alpha\epsilon_\tau} \sqrt{\frac{1}{2N} \log\frac{2|\mathcal{S}||\mathcal{A}|}{\delta}}
      \\
      & \le \frac{2}{f(z_\tau)} \sqrt{\frac{1}{2N} \log\frac{2|\mathcal{S}||\mathcal{A}|}{\delta}}
      & \mbox{by \eqref{eq:epsilon-UB}}
      \\
      & = 2 \Delta(s,a,z_\tau),
  \end{align*}
  where the first inequality holds since $f$ is $\alpha$-Lipshitz,
  and $f(\bar z_\tau)$ is $\epsilon_\tau$-neighbor of $f(z_\tau)$.
  This establishes \eqref{eq:pointwise_concentration_of_DDE_specific}.
  \prooflinkfinal{density_aware_concentration_of_ede}
\end{proof}

\subsection{Proof of Theorem \ref{claim:fixed_point_of_dde}}

\begin{proof}
\label{proof:fixed_point_of_dde}

  For brevity, let us write
  $\calT^\pi$ and $\calQ_\phi$ in terms of CDF and quantile, respectively.
  In addition, let us denote $\calI$ and $\calI^{-1}$ the inverse operator on quantile and CDF, respectively.
  Specifically,
  we write
  \begin{align*}
      \calT^\pi F(x;s,a)
      & = \sum_{s',a'} p^\pi(s',a'|s,a) \int F\left(\frac{x-r}{\gamma};s',a'\right) \, {\rm d}F_{\mathcal{R}(s,a)}(r), \\
      \calQ_\phi F^{-1}(\tau;s,a)
      & = F^{-1}(\tau;s,a) - \phi(s,a,\tau), \\
      \calI^{-1} F(\tau;s,a)
      & = F^{-1}(\tau;s,a) = \inf\left\{x\mid F(x;s,a)\ge\tau\right\}, \quad\text{and}\\
      \calI F^{-1}(x;s,a)
      & = F(x;s,a).
  \end{align*}
  Thus, the distorted Bellman operator on quantile can be written as
      $\calQ_\phi \calI^{-1} \calT^\pi \calI$.
  Let
  $\overline{\phi} \coloneqq \sup_{s,a,\tau} \phi(s,a,\tau)$
  and
  $\underline{\phi} \coloneqq \inf_{s,a,\tau} \phi(s,a,\tau)$.
  For any $s,a,x$, we have
  \begin{align}
      F(x+\underline\phi;s,a)
      \le \calI\calQ_\phi\calI^{-1}F(x;s,a)
      \le F(x+\overline\phi;s,a).
      \label{eq:bound1}
  \end{align}

  For $c\in\{\underline{\phi},\overline{\phi}\}$, let $F^{(c)}(x)\equiv F(x+c;s,a)$.  Then
  \begin{align}
      & \calI^{-1}\calT^\pi F^{(c)}(\tau)
      \nonumber\\
      & = \inf\left\{x\mid
      \sum_{s',a'} p^\pi(s',a'|s,a) \int F^{(c)}\left(\frac{x-r}{\gamma};s',a'\right) \, {\rm d}F_{\mathcal{R}(s,a)}(r) \ge \tau
      \right\}
      \nonumber\\
      & = \inf\left\{x\mid
      \sum_{s',a'} p^\pi(s',a'|s,a) \int F\left(\frac{x+\gamma\,c-r}{\gamma};s',a'\right) \, {\rm d}F_{\mathcal{R}(s,a)}(r) \ge \tau
      \right\}
      \nonumber\\
      & = \inf\left\{z\mid
      \sum_{s',a'} p^\pi(s',a'|s,a) \int F\left(\frac{z-r}{\gamma};s',a'\right) \, {\rm d}F_{\mathcal{R}(s,a)}(r) \ge \tau
      \right\} - \gamma\,c
      \nonumber\\
      & = \calI^{-1}\calT^\pi F(\tau;s,a) - \gamma\,c
      .
      \label{eq:shift}
  \end{align}
  Suppose that we update $F_t^{-1}$ with undistorted Bellman operator $\calB$ and $\underline F_t^{-1}$ with distorted Bellman operator $\calB^\phi$ from $F_0^{-1}=\underline F_0^{-1}$:
  \begin{align*}
      F_{t+1}^{-1}
      & = \calB F_t^{-1}
      = \calI^{-1} \calT^\pi \calI F_t^{-1}
      ,\quad \text{and} \quad
      \underline F_{t+1}^{-1}
      & = \calB^\phi \underline F_t^{-1}
      = \calQ_\phi \calI^{-1} \calT^\pi \calI \underline F_t^{-1}
      .
  \end{align*}
  We will show
  \begin{align}
      F^{-1}_t(\tau;s,a) - \sum_{k=1}^t \gamma^k \overline\phi
      \le \underline F^{-1}_t(\tau;s,a)
      \le F^{-1}_t(\tau;s,a) - \sum_{k=1}^t \gamma^k \underline\phi, \forall\tau
      \nonumber
  \intertext{or equivalently}
      \calI^{-1} F^{\overline\phi_k}_t(\tau;s,a)
      \le \calI^{-1}\underline F_t(\tau;s,a)
      \le \calI^{-1} F^{\underline\phi_k}_t(\tau;s,a), \forall\tau,
      \label{eq:induction}
  \end{align}
  where
  $\overline\phi_k \coloneqq \sum_{k=1}^t \gamma^{k-1} \,  \overline\phi$
  and
  $\underline\phi_k \coloneqq \sum_{k=1}^t \gamma^{k-1} \, \underline\phi$.
  When $t=0$, we have $F_0^{-1}=\underline F_0^{-1}$, thus \eqref{eq:induction} holds with equalities.  Suppose that \eqref{eq:induction} holds at $t$.  Then,
  since
  \begin{align}
      F(x)\le G(x), \forall x
      & \Rightarrow \calT^\pi F(x)\le \calT^\pi G(x), \forall x
      \label{eq:T}\\
      F(x)\le G(x), \forall x
      & \Rightarrow \calI^{-1} F(\tau)\ge \calI^{-1} G(\tau), \forall \tau \label{eq:I}\\
      F^{-1}(\tau)\le G^{-1}(\tau), \forall \tau
      & \Rightarrow \calQ^{\phi} F^{-1}(\tau)\le \calQ^{\phi} G^{-1}(\tau), \forall \tau,
      \label{eq:Q}
  \end{align}
  we have
  \begin{align}
      \underline F_{t+1}^{-1}(\tau;s,a)
      & = \calQ_\phi \calI^{-1} \calT^\pi \calI \underline F_t^{-1}(\tau;s,a)
      \nonumber\\
      & = \calQ_\phi \calI^{-1} \calT^\pi \calI \calI^{-1} \underline F_t(\tau;s,a)
      \nonumber\\
      & \ge \calQ_\phi \calI^{-1} \calT^\pi \calI \calI^{-1} F_t^{\overline\phi_t}(\tau;s,a)
      \nonumber\\
      & = \calQ_\phi \calI^{-1} \calT^\pi F_t^{\overline\phi_t}(\tau;s,a)
      \nonumber\\
      & = \calQ_\phi \calI^{-1} \calT^\pi F_t(\tau;s,a) - \gamma \, \overline\phi_t
      \label{eq:apply_shift}\\
      & = \calI^{-1} \calT^\pi F_t(\tau;s,a) - \phi(s,a,\tau) - \gamma \, \overline\phi_t
      \label{eq:refine}\\
      & \ge \calI^{-1} \calT^\pi F_t(\tau;s,a) - \overline\phi - \gamma \, \overline\phi_t
      \nonumber\\
      & = \calI^{-1} \calT^\pi F_t(\tau;s,a) - \overline\phi_{t+1},
      \nonumber
  \end{align}
  where the first inequality follows from the inductive hypothesis \eqref{eq:induction} and \eqref{eq:T}-\eqref{eq:Q}, and \eqref{eq:apply_shift} follows from \eqref{eq:shift}.
  Analogously,
  \begin{align*}
      \underline F_{t+1}^{-1}(\tau;s,a)
      & \le \calI^{-1} \calT^\pi F_t(\tau;s,a) - \underline\phi_{t+1}.
  \end{align*}
  These establish \eqref{eq:induction} for $t\gets t+1$.

  Letting $t\to\infty$, we obtain
  \begin{align*}
      F^{-1}_\infty(\tau;s,a) - \frac{\overline\phi}{1-\gamma}
      \le \underline F^{-1}_\infty(\tau;s,a)
      \le F^{-1}_\infty(\tau;s,a) - \frac{\underline\phi}{1-\gamma}, \forall\tau
      .
  \end{align*}

  The expression \eqref{eq:refine} and an analogous upper bound give a refinement:
  \begin{align*}
      F^{-1}_{t+1}(\tau;s,a) - \phi(s,a,\tau) - \gamma \, \overline\phi_t
      &\le \underline F_{t+1}^{-1}(\tau;s,a)
      \\
      &\le F^{-1}_{t+1}(\tau;s,a) - \phi(s,a,\tau) - \gamma \, \underline\phi_t
      \\
  \intertext{and letting $t\to\infty$:}
      F^{-1}_\infty(\tau;s,a) - \phi(s,a,\tau) - \frac{\gamma}{1-\gamma} \, \overline\phi
      &\le \underline F_\infty^{-1}(\tau;s,a)
      \\
      &\le F^{-1}_\infty(\tau;s,a) - \phi(s,a,\tau) - \frac{\gamma}{1-\gamma} \, \underline\phi
      .
  \end{align*}
\prooflinkfinal{fixed_point_of_dde}
\end{proof}

\end{document}